\documentclass{article}

\pdfoutput=1
\PassOptionsToPackage{numbers, compress}{natbib}
\usepackage[preprint]{neurips_2023}

\usepackage[utf8]{inputenc} % allow utf-8 input
\usepackage[T1]{fontenc}    % use 8-bit T1 fonts
\usepackage{hyperref}       % hyperlinks
\usepackage{url}            % simple URL typesetting
\usepackage{booktabs}       % professional-quality tables
\usepackage{amsfonts}       % blackboard math symbols
\usepackage{nicefrac}       % compact symbols for 1/2, etc.
\usepackage{microtype}      % microtypography
\usepackage{xcolor}         % colors
\usepackage{enumitem}
\usepackage{amsmath}
\usepackage{graphicx}
\usepackage{multirow}
\usepackage{makecell}
\usepackage{caption}

\graphicspath{{/}} %图片在当前目录下的figures目录

\title{CDIDN: A Registration Model with High Deformation Impedance Capability for Long-Term Tracking of Pulmonary Lesion Dynamics}

\author{%
  Xinyu Zhao \\
  College of Software \\
  Jilin University, CN \\
  \texttt{xinyuz9920@gmail.com} \\
   \And
   Sa Huang \\
   Department of Radiology\\
   Second Hospital of Jilin University \\
   \texttt{huangsa@jlu.edu.cn} \\
   \AND
  Wei Pang \\
  School of Mathematical and Computer Sciences \\
  Heriot-Watt University, UK \\
  \texttt{w.pang@hw.ac.uk} \\
   \And
   You Zhou\thanks{Corresponding author: zyou@jlu.edu.cn} \\
   College of Computer Sciences \\
   Jilin University, CN \\
   \texttt{zyou@jlu.edu.cn} \\
}

\begin{document}

\maketitle

\begin{abstract}
We study the problem of registration for medical CT images from a novel perspective---the sensitivity to degree of deformations in CT images. Although some learning-based methods have shown success in terms of average accuracy, their ability to handle regions with local large deformation (LLD) may significantly decrease compared to dealing with regions with minor deformation. This motivates our research into this issue. Two main causes of LLDs are organ motion and changes in tissue structure (such as lesions), with the latter often being a long-term process. Traditional methods lack real-time capability due to their time-consuming nature. In this paper, we propose a novel registration model called Cascade-Dilation Inter-Layer Differential Network (CDIDN), which exhibits both high deformation impedance capability (DIC) and accuracy. CDIDN improves its resilience to LLDs in CT images by enhancing LLDs in the displacement field (DF). It uses a feature-based progressive decomposition of LLDs, blending feature flows of different levels into a main flow in a top-down manner. It leverages Inter-Layer Differential Module (IDM) at each level to locally refine the main flow and globally smooth the feature flow, and also integrates feature velocity fields that can effectively handle feature deformations of various degrees. We assess CDIDN using lungs as representative organs with large deformation. Our findings show that IDM significantly enhances LLDs of the DF, by which improves the DIC and accuracy of the model. Compared with other outstanding learning-based methods, CDIDN exhibits the best DIC and excellent accuracy. Based on vessel enhancement and enhanced LLDs of the DF, we propose a novel method to accurately track the appearance, disappearance, enlargement, and shrinkage of pulmonary lesions, which effectively addresses detection of early lesions (1 mm or larger) and peripheral lung lesions, issues of false enlargement, false shrinkage, and mutilation of lesions.
\end{abstract}

\section{Introduction}
%ok,总述深度学习在配准中的作用，引出配准对医疗的作用
The application of deep learning in medical image processing has made significant progress, enabling more accurate disease diagnosis \cite{singh20203d,ker2019image,singh2020shallow,ker2017deep}. For example, Esteva et al. has successfully applied deep learning to diagnose cancer \cite{esteva2017dermatologist}, while Gulshan et al. has developed an automated system to detect diabetic retinopathy and macular edema \cite{gulshan2016development}. Medical image registration based on deep learning has the potential to address numerous complex medical challenges, making it a hot topic of research in the field. Therefore, deep learning-based registration techniques for medical images are increasingly gaining attention.

%ok,指出现在模型中存在的问题，提出我们的目标
Although some current methods have achieved decent average accuracy on static and dynamic organs \cite{kearney2018unsupervised, de2017end,sentker2018gdl,eppenhof2018pulmonary,fu2019automatic, fu2020lungregnet}, their stability often proves insufficient due to sensitivity to deformation degree in organs and tissues. Regions with local large deformation (LLD) require more attention, as they may indicate areas of lesion changes or significant organ motion. Precise alignment of LLDs is crucial for registration-based medical diagnosis. Thus, while aiming for high average accuracy, we seek stability in model accuracy despite various local deformation degrees in medical images. To achieve this, we introduce the concept of deformation impedance capability (DIC), representing a model's ability to resist accuracy decrease as local deformation degrees increase in medical images.

%ok, 引出用肺部作为大形变代表器官进行配准和评估的原因
Lung image registration poses significant challenges due to lung deformability. Lungs exhibit various degrees of both global and local deformation, with greater local deformation closer to the edges. Even separating lung CT images of the same patient with a large time interval can still show LLDs as there exist changes in tissue structure and potential lesions. Despite the extensive research on deformable image registration (DIR) in lung images \cite{sotiras2013deformable,haskins2020deep}, both global and local deformations still pose challenges to the DIC, accuracy, and speed of registration methods.

%ok
Through the above analysis, we have conducted extensive research using lungs as representative organs with both large global and local deformations. Our work makes the following contributions: 

%ok
\begin{enumerate}[label=(\arabic*)]
	\item We propose CDIDN, an unsupervised end-to-end registration model with low sensitivity to both global and local deformations and high average accuracy. CDIDN exhibits excellent registration accuracy even in areas with LLD, demonstrating high DIC.
	
	\item We propose Adaptive Regulation U-Net (ARU), which adjusts the depth and skip connections of encoder-decoder layers according to feature size and level. It generates a set of appropriate feature displacement field (FDF) prediction networks for multiscale features and avoids excessive coverage of receptive fields (RFs) while saving memory and computational time.
	
	\item We propose the Inter-Layer Differential Module (IDM) that significantly enhances LLDs in the displacement field (DF), improving the model's DIC by improving its resilience to deformations. IDM can be applied to feature- or cascade-based methods, including CDIDN.
	
	\item We propose a novel method for long-term tracking of pulmonary lesion dynamics by vessel enhancement and enhanced LLDs of the DF. It utilizes a formula visualization technique, effectively addressing some medical challenges including detection of early lesions and peripheral lung lesions, false enlargement, shrinkage, and mutilation of lesions.
\end{enumerate}

\section{Related Works}
\label{gen_inst}

\paragraph{Registration based on deep learning and small deformation static displacement field.} 

%ok
Deep learning-based DIR methods have been proven to be effective in analysing various types of medical images, like MRI brain \cite{wu2015scalable}, CT head and neck \cite{kearney2018unsupervised}, CT thorax \cite{jiang2018cnn}, MR/US prostate \cite{hu2018weakly}, and 4D-CT lung images \cite{de2017end,sentker2018gdl,eppenhof2018pulmonary,fu2019automatic}. Eppenhof et al. used a supervised CNN with U-Net \cite{eppenhof2018pulmonary} and trained it using synthetic random transformations. However, that study has limitations as the random transformations may differ significantly from the actual lung motion, and the supervised training with random transformations may not provide effective regularization for the DF. Jaderberg et al. proposed the Spatial Transformer Networks (STN) \cite{jaderberg2015spatial}. Inspired by this, some researchers added an STN layer after CNN to obtain the warped image using the DF for end-to-end image registration \cite{zhao2019unsupervised, balakrishnan2018unsupervised, balakrishnan2019voxelmorph}. However, supervised learning-based methods need excessive labeled data and has limited performance, hindering practical applicability. To address this, unsupervised methods have been proposed. VoxelMorph \cite{balakrishnan2018unsupervised} and VTN \cite{zhao2019unsupervised} use deconvolution layers \cite{noh2015learning} to predict dense, high-performing DFs. VoxelMorph was evaluated on brain MRI scans \cite{balakrishnan2018unsupervised,balakrishnan2019voxelmorph}, but it later demonstrated limitations on other datasets \cite{zhao2019unsupervised}. VTN uses an initial CNN to execute affine transformation before generating DFs, establishing a genuine end-to-end registration method. However, these methods produce static DFs only applicable for registering static organs or tissues with minor deformations, and the effect of LLDs and real-time organ motion on registration is disregarded.

%ok
\paragraph{Hierarchical registration based on recursive and cascaded methods.} Lin et al. proposed Feature Pyramid Networks (FPN) \cite{lin2017feature}, which enriches all scale features with semantic information through top-down lateral connections. Prior to FPN, most registration algorithms relied only on the original CT images.  While this model considers registration at both the aspects of original images and features, it does not account for the transmission of deformations between different levels, and it still uses a static DF method to describe the LLDs caused by organ motion. Recursive Cascaded Networks \cite{zhao2019recursive} refines the DF using identical basic network blocks and cumulative registration. However, it still uses a static DF method to describe LLDs as in FPN. Our method employs feature-level cascading instead of image-level cascading, refining a main flow with the FDF at each level, propagating the warped feature to a lower level by a learnable deconvolution kernel. This preserves higher semantic of the higher-level features and higher accuracy for target location prediction.

%ok
\paragraph{Registration based on velocity field.} Dalca et al. further improved VoxelMorph \cite{dalca2019unsupervised} by proposing a probabilistic generative model and deriving an unsupervised learning-based inference algorithm. Their approach transforms the traditional static DF into a velocity field based on diffeomorphism, assigning "speed" to the displacement vector of each voxel. This better characterizes the actual motion patterns of organs and tissues in large deformation organ registration, resulting in velocity fields that have been integrated by an initial DF. However, making a single prediction on LLDs of CT images based solely on the raw image data is insufficient, which is showed in our comparative experiments. In our study, we utilizes a feature-based, progressive decomposition of LLDs and performed gradually decreasing "velocitization" operations on the FDFs from low level to high level. The experimental results indicate that applying different degrees of "velocitization" to the feature levels can enhance the model's DIC and accuracy.

\section{Method}
\label{headings}

%ok
\subsection{Lung Parenchyma Segmentation (LPS) and Pulmonary Vessel Enhancement (PVE)}

\begin{figure}[!htbp]
	\centering
	\includegraphics[width=1\textwidth]{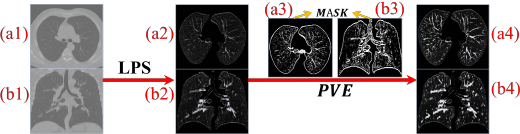} % 插入图片
	\caption{Pre-processing. (a1, b1): original images. (a2, b2): images after LPS. (a3, b3): normalized results after 2D Frangi filtering on tri-orthogonal planes. (a4, b4): images after PVE ($E_p^N$). }
	\label{p1}
\end{figure}

To enhance CT registration accuracy, pre-processing is necessary to mitigate noise impact on CT backgrounds. We utilize binary clustering for LPS with an intensity window of [-1000,700] to avoid extreme intensities influencing the clustering centers. We set minimum retention areas for each lung slice to be 300 voxels to eliminate peripheral lung and tracheal tissues. For PVE, we perform 2D Frangi filtering on tri-orthogonal planes \cite{frangi1998multiscale}. However, relying solely on 2D Frangi filtering introduces much white noise, mainly corresponding to intensities in [-1000,-800], which are not vessels or lesions. Meanwhile, to increase enhancement magnitude for low-intensity voxels, we consider using a logarithmic function. In summary, we input each voxel of CT images into the following equation:
\begin{equation}
	E_p = \begin{cases}
	I_p,  &\text{if } I_p \leq B[0] + D;\\
	I_p + (B[1] - I_p) \times \ln((e-1) \times M + 1), &\text{if } I_p > B[0] + D.
	\end{cases},
	\label{f1}
\end{equation}
where $E_p$ and $I_p$ denote the intensity of the voxel $p$ after and before PVE respectively, $B$ denotes intensity window, and $D$ denotes the distance between the lower limit of the enhanced intensities and CT backgrounds (i.e., $B[0]$), $M$ denotes the Mask as shown in Figure~\ref{p1} (a3) and (b3). In this experiment, we set $D$ to be 200. We enhance voxels in CT images using equation~\eqref{f1} and then process them using the normalization method as follows:
\begin{equation}
E_p^N = \frac{E_p-B[0]}{B[1] - B[0]}.
\label{f2}
\end{equation}

Figure~\ref{p1} shows pre-processing process of lung CT images. To further optimize the image quality, we center and crop the lung parenchyma portion of CT images, as it is the area of our focus.

\subsection{Cascade-Dilation Inter-Layer Differential Network (CDIDN)}

\begin{figure}[!htbp]
	\centering
	\includegraphics[width=1\textwidth]{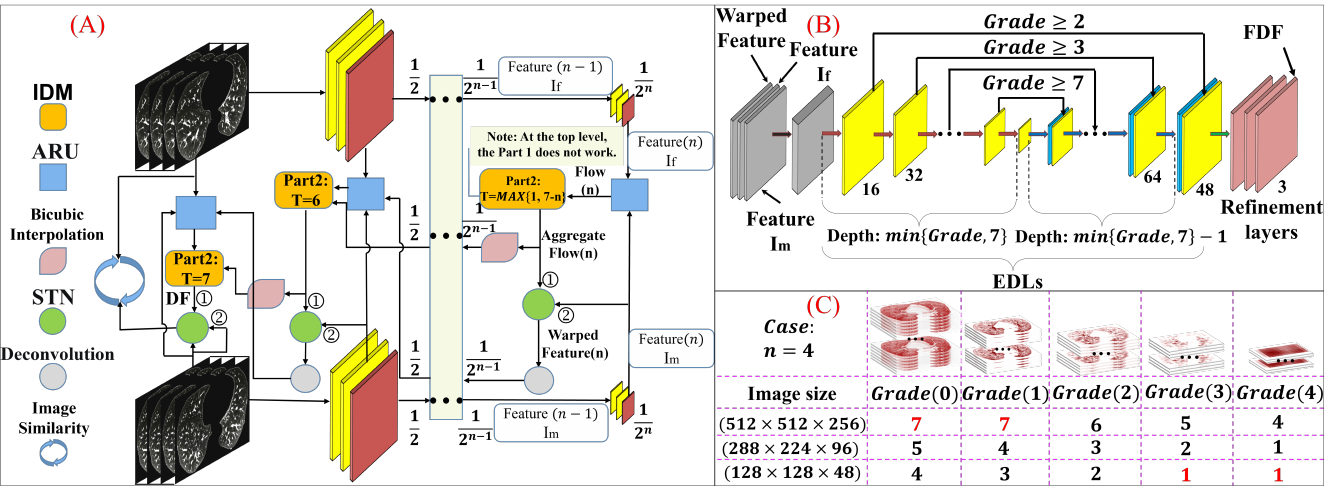} % 插入图片
	\caption{(A)The architecture of CDIDN. (B) The architecture of ARU. Its depth at each level is limited by controlling depth of encoding-decoding layers. (C): A case for ARU when $n=4$ .}
	\label{p2}
\end{figure}

%ok
CDIDN (Figure~\ref{p2} (A)) adopts a dual-stream feature pyramid and a top-down manner to generate FDFs and Aggregate Flow, with three convolutions per level to extract features. It aggregates the Flow from each level into a main flow (Aggregate Flow), which locally refines the Aggregate Flow with Flows and globally smooths the Flow at each level. By progressively decomposing LLDs into features and registering them at each level, CDIDN reduces the impact of LLDs, preserving high-level semantics and enhancing accuracy. The Aggregate Flow is dilated by bicubic interpolation to be  smoother and more accurate, and the advantage of bicubic interpolation is supported by experimental results \cite{fu2020lungregnet}.

\subsubsection{Adaptive Regulation U-Net (ARU)}

 Multiple convolution operations are required in feature extraction. The amount of data in features decreases as the level increases. If a conventional U-Net is used to process high-level features and predict FDFs, it will result in features passing through a deep network, with a fewer-layer network actually having an equivalent effect. We call this issue "RF over-coverage" problem, which causes feature points of FDFs at high levels to become similar at the beginning of training, making the model difficult to train and a slower registration speed. When processing CT images and their features, depth denotes the number of slices, which is usually less than the width and height of slices. Let $\phi_{ED}^i$ be the output of the encoding-decoding layers (EDLs) in ARU at level $i$. We must ensure that at least one high-level $\phi_{ED}^i$ can consider all slices of $Feature(i)$, i.e., the RF in the depth direction covers all slices, to ensure that the process of refinement maintains globality in the Aggregate Flow. The ARU can adjust the depth and number of skip connections of the EDLs based on feature level and size. This generates an appropriate FDF prediction network, which ensures that the RF for each point of the FDF does not exceed the depth of features at each level. This approach avoids the "RF over-coverage" problem. Figure~\ref{p2} (B) depicts the architecture of ADU, with $I_f^i$, $I_m^i$ and $W^i$  representing the fixed, moving and warped features at level $i$ respectively. Furthermore, $I_f^0$, $I_m^0$, and $W^0$ correspond to the fixed, moving, and warped CT image, respectively. The ARU takes as input $I_f^i$, $I_m^i$, and $W^{i+1}$, and outputs the FDF of level $i$. Prior to running the model, the ARU calculates the Grade for each level using equation as follows:
\begin{equation}
	Grade(i) = \begin{cases}
		\min\{\max\{D^i, 1\}, \,7\},  &i=n;\\
		\min\{Grade(i+1)+1 ,\,\max\{D^i, 1\},\, 7\}, &0\leq i < n.
	\end{cases},
	\label{f3}
\end{equation}
where $D^i$ calculated using RF is the maximum depth of the EDLs that satisfies the condition that the RF of $\phi_{ED}^i$ does not exceed the number of slices of the features at level $i$. Then, the corresponding FDF prediction network structure for each level based on its Grade can be generated.

Figure~\ref{p2} (C) shows the impact of CT image size on the Grades of different levels when $n=4$. In theory, for any combination of $n$ and CT image size,  CDIDN can generate a set of ARUs with appropriate depths. Specifically, for a single ARU, as the Grade increases by 1, the encoder and decoder of ARU will add Conv 1, a max pooling layer, Conv 2, an upsampling layer successively at the junction, with a skip connection being added.

\subsubsection{Inter-Layer Differential Module (IDM)}

\begin{figure}[!htbp]
	\centering
	\includegraphics[width=1.0\textwidth]{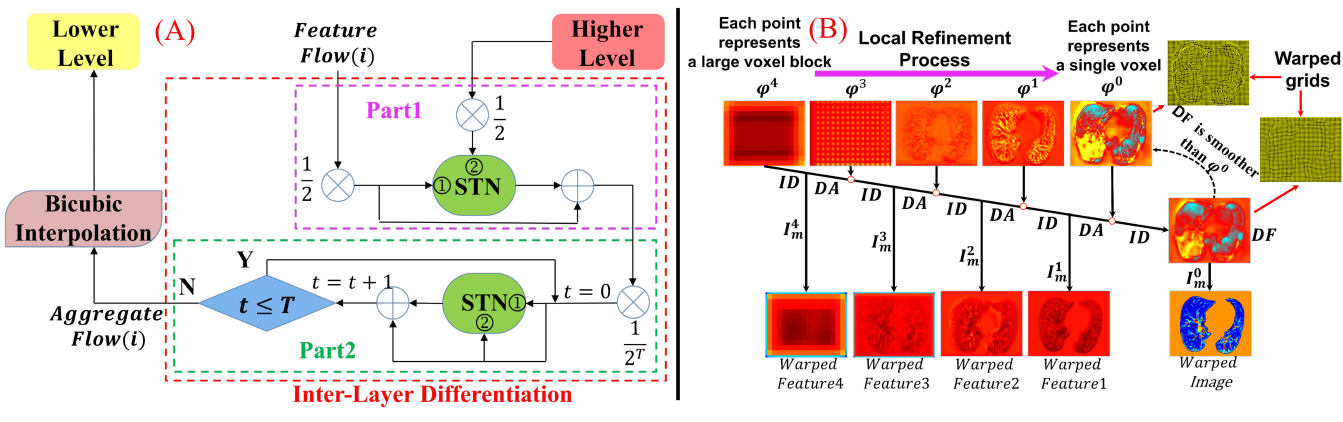} % 插入图片
	\caption{IDM. (A): Architecture. The input end 1 and input end 2 of STN represent its action end and acted end, respectively. (B) IDM's local refinement and global smoothing effects at each level.}
	\label{p3}
\end{figure}

IDM (Figure~\ref{p3} (A)) is used to refine (F)DFs to accurately describe voxel movements in CT images. It involves two main parts. Part 1 takes two inputs: the Aggregate Flow from a higher level and the Flow at the current level. We use these as inputs to the STN, with the Flow acting as the DF. We achieve local refinement and global smoothing by resampling and connection. Each feature point in the high-level's Aggregate Flow represents the overall movement direction of a large voxel block of in the moving image. By passing the Aggregate Flow down level by level, the Flow obtained from lower-level features can be continuously used to refine the Aggregate Flow locally, transforming from large voxel blocks to smaller ones. By cascading these operations across all levels, we can achieves refined granularity of the Aggregate Flow to match the displacement of each voxel in the CT image.

Lower-level features have a greater impact from LLDs, leading to inaccurate motion descriptions by static FDFs. This implies that static final FDFs may not exist. Inspired by VoxelMorph \cite{dalca2019unsupervised} and improved upon it, Part 2 uses the static FDF as an initial feature velocity field (FVF), which is integrated to obtain the deformation field (DMF) for each level. The DMF ($\phi_t^j$) is defined as follows:
\begin{equation}
\frac{\partial\phi_t^j}{\partial t }  =v(\phi_t^j )=v_t^j \circ \phi_t^j,
\label{f4}
\end{equation}
where $j$ is the feature level, $t$ is time and $t \in [0,1]$, $v_t^j$ is the FVF, $\phi_0^j$ is an identity mapping, and $\circ$ is a composition operator. If we have $\phi_{1/2^k}^j$ and $v_k^{j}$ ($k \in [1,T]$), we have $\phi_{1/2^{k-1}}^j=\phi_{1/2^k}^j+v_k^{j} \circ \phi_{1/2^k}^j$, $v_{k-1}^j = v_{k}^j+v_{k}^j \circ \phi_{1/2^k}^j$. From $\phi_{1/2^T}^j={\phi_0^j}+v(\phi_0^j)={\phi_0^j}+v_T^j$, we can recursively derive $\phi_1^j$ and $v_0^j$. A suitable $T$ should be determined based on the level and be chosen so that the initial FVF at level $j$ satisfies $v_T^j \approx 0$. Based on experience, we set $T=7$ for level 0. As the level increases, the feature dimension is halved and the effect of original large deformations weakens, so we progressively decrease $T$ from level 0 to $n$. Using the output from Part 1 of IDM as the initial static FVF ($v_T^j$) and integrating it in a manner of $T$ iterations, we get the final FDF ($v_0^j$) and DMF ($\phi_1^j$) for this level. Finally, $\phi_1^j$ is used to warp $I_m^j$, and $v_0^j$ is used as the Aggregate Flow of level $j$ and subjected to bicubic interpolation, with the output being propagated to level $(j-1)$ for further processing.

Let ${AF}^i$ be the Aggregate Flow at level $i$, $DA(\cdot)$ be the dilation function, $ID(\cdot)$ be the IDM function, and $F^i(\cdot)$ be the ARU function at level $i$. We cascade the entire registration process by recursively performing registration on each level's features and the warped feature from the higher level. The final result, $W^0$, is the composite of the DF and the moving CT image, namely
\begin{equation}
	W^0=DF \circ I_m^0.
	\label{f5}
\end{equation}
Theoretically,
\begin{equation}
	{AF}^i = \begin{cases}
	{ID}(F^n(I_f^n,\,I_m^n)),  &i=n;\\
	{ID}({DA}({AF}^{i+1} ) \circ F^i({DA}({AF}^{i+1} \circ {I_m}^{i+1}),\,I_f^i,\, I_m^i)), &0\leq i < n.
	\end{cases}.
\label{f6}
\end{equation}
Let $\phi^i$ denote the Flow generated using the function $F(\cdot)$ at level $i$, then we have:
\begin{equation}
	DF ={AF}^0=ID(\phi^0 \circ {DA}({ID}(\phi^1 \circ \dots \circ {DA}({ID}(\phi^{n-1} \circ {DA}({ID}(\phi^n)))) \dots ))).
	\label{f7}
\end{equation}

As shown in Figure~\ref{p3} (B), the features of the moving image are progressively warped from the highest to the lowest level through the recursive process of this cascade dilation. Through this level-by-level local refinement and global smoothing, the DF obtained is smoother and more realistic compared to directly obtaining it from $I_f^0$, $I_m^0$, and $W^1$. For example, it is observed that the smoothness of the DF is stronger than that of $\phi^0$, and the former more accurately reflects the real motion of the lungs.

\subsection{Loss Function}
The loss function consists of similarity loss and regularization loss, with the latter being used to prevent unrealistic deformations of the DF. The similarity loss is measured by NCC, and the regularization loss is represented by the spatial gradient of the DF. The difference between adjacent voxels is used as an approximation for the spatial gradient.The loss function $\mathcal{L}(\cdot, \cdot, \cdot)$ is defined as follows:
\begin{equation}
\mathcal{L}(I_f^0,\,  I_m^0,\, {DF}) = -NCC(I_f^0, \, W^0) + \lambda\sum_{p \in \Omega}\| \nabla DF(p)\|^2,
\label{f8}
\end{equation}
where $\nabla DF(p)$ denotes the displacement vector at voxel $p$ and $\lambda$ is regularization coefficient.
\section{Experiments}

\subsection{Experimental Setup}
In this section, we train four CDIDN variants and compare CDIDN with other outstanding learning-based registration methods. We evaluate three aspects: the efficacy of IDM's Part 1 and Part 2 on registration accuracy, IDM's impact on DIC, and CDIDN's DIC compared to other methods. 

\paragraph{Data}

We use three datasets, namely the POPI dataset \cite{vandemeulebroucke2007popi}, the Dir-lab dataset \cite{castillo2009four,castillo2009framework}, and a dataset collected from our institution. We use the POPI dataset and our institution's dataset for training. The POPI dataset includes 6 cases, each of which records a set of 10 3D-CT lung images for a respiratory cycle. Our institution's dataset includes 42 cases, each of which includes 2-4 CT images, and 40 cases are selected for training. Unlike the Dir-lab and POPI datasets, our institution's CT images of the same patient were taken with different intervals, ranging from days to months, which may lead to significant lesion changes that can be regarded as special cases of LLD registration. Training with data possessing such characteristics can enhance the model's generalization and robustness. For testing, we use the Dir-lab dataset consisting of 10 cases of 4D-CT, each with 300 expert landmark pairs (ELPs) labeled on T00 (end-inspiratory) and T50 (end-expiratory).  All lung CT images were segmented and vessel-enhanced before training, with size of [288, 224, 96] and spacing of [1, 1, 2.5].

\paragraph{Model}
We set the maximum feature level to be 4 and $\lambda$ to be 0.5. During training, we perform all permutations and combinations of any two CT images from all 3D-CTs contained in each 4D-CT dataset or each patient's dataset, and use all image pairs for training. We adopt the Adam optimizer with an initial learning rate of 1e-4, and use the cosine annealing algorithm to dynamically adjust the learning rate, gradually decreasing it from 1e-4 to 1e-5 throughout the entire training process. We train CDIDN for 1500 epochs using five-fold cross-validation with the batch size of 2. Each model is undergoing parallel training on three NVIDIA RTX 3090 GPUs (72GB memory).

\subsection{Results}

\subsubsection{The Impact of Part 1 and Part 2 of IDM on Registration Accuracy}

We compare four variants of CDIDN: CDIDN-v1 (with simple connection), CDIDN-v2 (with Part 1 but not Part 2), CDIDN-v3 (with simple connection and Part2), and CDIDN-v4 (with Part 1 and Part 2). We compute the target registration error (TRE) by registering T50 to T00 within each case. Table~\ref{t1} demonstrates the TRE results for CDIDN-v1 to CDIDN-v4, which shows improved accuracy across the four variants and CDIDN-v4 has the best performance. We also conduct a set of experiments for CDIDN-v4 using data without PVE. The result indicates that PVE helps improve accuracy significantly. Based on the conducted experiments, it is evident that both Part 1 and Part 2 of IDM contribute to improving the registration performance. Part2 yields a greater improvement in the registration performance than Part 1, and their combined usage yields the optimal outcome.

%We evaluate their registration effects on T00 and T50 of Dir-lab dataset's CASE10 by comparing the intensity differences between T00 and the warped images from these variants. As shown in Figure~\ref{p4}, CDIDN-v4 has the smallest intensity differences and best performance.

%\begin{figure}[!htbp]
%	\centering
%	\includegraphics[width=0.5\textwidth]{pic/intensity_delta} % 插入图片
%	\caption{Intensity difference images. (a1, b1): between the fixed and moving images before registration, (a2–a5, b2–b5): between the fixed and warped images from CDIDN-v1 to CDIDN-v4.}
%	\label{p4}
%\end{figure}

\begin{table}[!htbp]

		\centering
		\caption{Target registration error values for different deep learning-based methods on Dir-lab datasets.}
		\resizebox{1\linewidth}{!}{
		\begin{tabular}{ccccccccc|c|cccc}

		\toprule
		\multirow{2}{*}[-1em]{\centering Dataset} & \multirow{2}{*}[-1em]{\centering \makecell[c]{TRE before\\ registration}}  & \multicolumn{12}{c}{\textbf{Deep learning-based methods}} \\
		\cmidrule(r){3-14}
		& & VoxelMorph \cite{dalca2019unsupervised} &
		\makecell[c]{De Vos et \\ al. \cite{de2017end}} &
		\makecell[c]{Sentker et \\ al. \cite{sentker2018gdl}} & 
		\makecell[c]{Eppenhof \\ et al. \cite{eppenhof2018pulmonary}} &
		\makecell[c]{ Sokooti et \\ al. \cite{sokooti20193d}} & 
		\makecell[c]{ Fechter et \\ al. \cite{fechter2020one}} & 
		\makecell[c]{ Jiang et \\ al. \cite{jiang2020multi}} & CDIDN-v4  & \makecell[c]{ CDIDN-v4 \\ (NO PVE)} & CDIDN-v1& CDIDN-v2& CDIDN-v3\\
		\cmidrule(r){1-14}
		1&	3.89 ± 2.78&	1.44 ± 0.75&	1.27 ± 1.16&	1.20 ± 0.60&	1.45 ± 1.06&	\textbf{1.13 ± 0.51}&	1.21 ± 0.88&	1.20 ± 0.63&	1.19 ± 0.60&	1.24 &1.44 ± 0.77&	1.32 ± 0.64&	1.21 ± 0.63\\
		2&	4.34 ± 3.90	&1.56 ± 0.78&	1.20 ± 1.12	&1.19 ± 0.63&	1.46 ± 0.76&	1.08 ± \textbf{0.55}&	1.13 ± 0.65& \textbf{1.13} ± 0.56	&1.32 ± 0.64&		1.33 &1.40 ± 0.68&	1.36 ± 0.65	&1.34 ± 0.66\\
		3&	6.94 ± 4.05	&1.92 ± 1.26&	1.48 ± 1.26&	1.67 ± 0.90&	1.57 ± 1.10&	1.33 ± 0.73&	1.32 ± 0.82&\textbf{	1.30 ± 0.70}&	1.39 ± 0.78&	1.43 &	1.42 ± 0.81&	1.42 ± 0.73&	1.42 ± 0.77\\
		4&	9.83 ± 4.86	&2.49 ± 1.58&	2.09 ± 1.93&	2.53 ± 2.01&	1.95 ± 1.32&	1.57 ± 0.99&	1.84 ± 1.76&	\textbf{1.55 ± 0.96}&	1.81 ± 1.10 &	1.83 &	1.91 ± 1.20&	1.82 ± 1.09	& 1.87 ± 1.11\\
		5&	7.48 ± 5.51&	2.30 ± 1.87	&1.95 ± 2.10&	2.06 ± 1.56&	2.07 ± 1.59&	\textbf{1.62} ± 1.30	&1.80 ± 1.60&	1.72 ± \textbf{1.28}&	1.82 ± 1.52 &1.86&	1.93 ± 1.74	&1.89 ± 1.50&	1.84 ± 1.50	\\
		6&	10.89 ± 6.97&	2.59 ± 1.66	&5.16 ± 7.09&	2.90 ± 1.70	&3.04 ± 2.73&	2.75 ± 2.91	&2.30 ± 3.78&	2.02 ± 1.70&	\textbf{1.65 ± 0.88}&1.87&	2.02 ± 1.38	&1.76 ± 0.98&	1.67 ± 0.90		\\
		7&	11.03 ± 7.43&	3.10 ± 2.25&	3.05 ± 3.01&	3.60 ± 2.99&	3.41 ± 2.75	&2.34 ± 2.32&	1.91 ± 1.56&	\textbf{1.70} ± 1.03&	1.76 ± \textbf{0.93}&	1.85&	1.95 ± 1.41&	1.83 ± 0.99&	1.80 ± 0.97\\
		8&	14.99 ± 9.01&	5.07 ± 3.39&	6.48 ± 5.37&	5.29 ± 5.52	&2.80 ± 2.46&	3.29 ± 4.32&	3.47 ± 5.00&	2.64 ± 2.78&	\textbf{2.53 ± 1.91}&	3.30&	3.34 ± 2.71&	2.97 ± 2.33&	2.57 ± 1.95\\
		9&	7.92 ± 3.98	&2.56 ± 1.50&	2.10 ± 1.66&	2.38 ± 1.46&	2.18 ± 1.24&	1.86 ± 1.47&	\textbf{1.47} ± 0.85	&1.51 ± 0.94&	1.58 ± \textbf{0.83}&	1.89&	1.70 ± 0.98&	1.73 ± 0.92&	1.60 ± 0.81\\
		10&	7.30 ± 6.35&	2.78 ± 2.61	&2.09 ± 2.24&	2.13 ± 1.88&	1.83 ± 1.36&	1.63 ± 1.29&	1.79 ± 2.24&	1.79 ± 1.61&	\textbf{1.59 ± 1.19}&	1.84&1.76 ± 1.27	&1.70 ± 1.17	&1.64 ± 1.24\\
		\cmidrule(r){1-14}
		Mean&	8.46 ± 6.55&	2.58 ± 2.15&	2.64 ± 4.32&	2.50 ± 1.16&	2.17 ± 1.89&	1.86 ± 2.12&	1.83 ± 2.35&	1.66 ± 1.44	&\textbf{1.66 ± 1.16}&1.84&	1.89 ± 1.52&	1.78 ± 1.28&	1.70 ± 1.18	\\
		\bottomrule
	\end{tabular}
	\label{t1}
}
\end{table}

\subsubsection{Evaluation of Model Accuracy and Deformation Impedance Capability}

\paragraph{Evaluation of CDIDN Model Accuracy.}
We train and test VoxelMorph \cite{dalca2019unsupervised} using the same method, and compare CDIDN-v4 with VoxelMorph and methods proposed by other scholars (Table~\ref{t1}), using TRE as metric. By comparison, our CDIDN-v4 achieves excellent registration accuracy.

\begin{figure}[!htbp]
	\centering
	\includegraphics[width=1\textwidth]{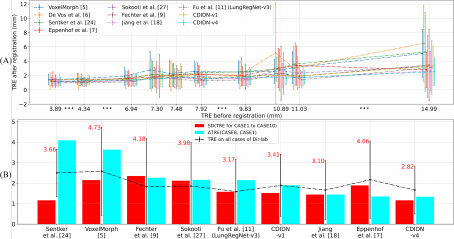} % 插入图片
	\caption{(A) Average TREs of different cases in Dir-lab before and after registration by different DIR methods. (B) Performance of different DIR methods on the Dir-lab dataset.}
	\label{p5}
\end{figure}

\paragraph{Evaluation of Deformation Impedance Capability of CDIDN.} When registering organs or tissues with LLDs, we prioritize stability, especially when accuracy is similar across various methods. The Dir-lab dataset has been standardized to neutralize the effect of distinct scanners and imaging parameters, and in each case, 300 ELPs are evenly dispersed throughout the lungs, and all 10 cases are non-lung disease patients. Given that we are actually registering ELPs and lung deformation induced by respiratory motion varies across individuals, the average TRE of the 300 ELPs before registration in each case can approximate the global lung deformation of each subject and identifies regions with significant LLDs by their TREs. We then evaluate the models' DIC from three perspectives.

\begin{enumerate}[label=(\arabic*)]
	\item \textbf{The impact of global deformation degree on registration accuracy.} Figure~\ref{p5} (A) shows average TREs on the Dir-lab dataset before and after registration using various methods. Although characteristics of methods and CT images may result in certain methods having larger individual TREs before registration but smaller TREs post-registration, overall, an increase in lung global deformation before registration corresponds to an increase in the average TRE after registration. We choose cases with the smallest (CASE1) and largest (CASE8) global deformation degrees and use the difference in their average TREs after registration as the metric, minimizing the effect of CT image characteristics or ELP selection strategy on model performance. Figure~\ref{p5} (A) and the blue bars in Figure~\ref{p5} (B) indicate that CDIDN-v4 has the slowest and smallest decrease in registration accuracy as the average TRE before registration increases significantly. Among the tested methods, CDIDN-v4 demonstrates least sensitivity to global deformation degree in terms of registration accuracy.

	\item \textbf{Sensitivity of registration ability to local deformation degree when global deformation degree is certain.} The standard deviation (SD) of TREs is a useful metric for evaluating model stability and robustness for each case, as it reflects the registration ability difference between large and small local deformation regions under a specific global deformation degree. Among all methods, CDIDN-v4 has the smallest SDs for cases with higher global deformation degree such as CASE6 to CASE10, and very small SDs for other cases. Overall, CDIDN-v4 demonstrates the lowest sensitivity to local deformation degree, with this advantage becoming more prominent as the global deformation degree increases significantly.
	
	\item \textbf{Sensitivity of registration ability to the degree of combined global and local deformation.} We use the SD of the TREs from all cases as the evaluation metric. Since the global deformation degree in the ten cases has a large span and each case has different degrees of local deformations, a smaller value of this metric implies lower sensitivity to the combined deformation degree. From the red bars in Figure~\ref{p5} (B), CDIDN-v4 has the smallest SD, indicating the highest DIC and registration stability under different scenarios.
	
\end{enumerate}

In summary, compared to outstanding methods proposed by Fu et al. \cite{fu2020lungregnet} (LungRegNet-v3, which is the state-of-the-art lung DIR method proposed in 2020) and other scholars, CDIDN-v4 shows the best DIC and best accuracy for LLDs. By the comparative analysis of CDIDN-v1 and CDIDN-v4 in Figure~\ref{p5} (A) and (B), we find that the IDM  has a significant effect on improving the model's DIC.

\subsubsection{CDIDN for Long-term Tracking of lesion dynamics}

\begin{figure}%[!htbp]
	\centering
	\includegraphics[width=1\textwidth]{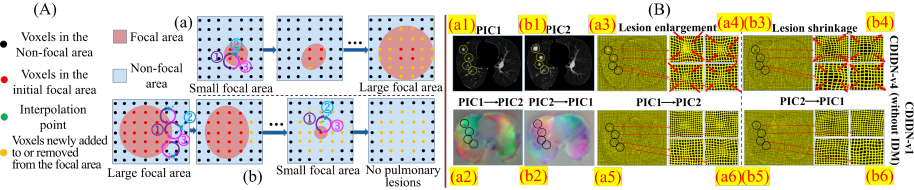} % 插入图片
	\caption{The role of the DF's LLDs in lesion changes. (A) CDIDN changes lesions during registration. (a) Enlargement. (b) Shrinkage and disappearance. Op1: Shift coordinates. Op2: Sampling. Op3: Bring back sampled values. (B) Lesion changes are reflected in DF and warped grids with LLDs. (a1, b1): CT images of a patient with lung disease at different time. (a2, b2): DF in two scenarios. (a3-a6, b3-b6): warped grids obtained by DF from CDIDN-v4 and  CDIDN-v1 with some major lesions.}
	\label{p6}
\end{figure}

Tracking dynamic lesion changes over extended periods can be thought as a LLD registration problem. Clinical scenarios frequently require longitudinal studies where multiple lung CT scans are taken several months apart, and lesions change a lot, which is different from 4D-CTs. Such studies require methods with high DIC to tackle LLDs at lesion locations. CDIDN-v4 generates DF with enhanced LLDs, enabling better monitoring lung lesion dynamics. Figure~\ref{p6} (A) shows how CDIDN enlarges, shrinks, or removes lesions. We select two lung CT scans (PIC1, PIC2) which are not as training samples in our institution's dataset, and they were obtained from a patient with a months-long interval.

 Figure~\ref{p6} (B) shows DFs and warped grids resulting from lesion enlargement and shrinkage. (a3)/(a4) and (b3)/(b4) confirm the content shown in Figure~\ref{p6} (A) that lesion locations experience LLDs. The warped grids tend to shrink towards the lesion center when enlargement and expand from the original lesion location when shrinkage. By comparing (a3-a4)/(b3-b4) with (a5-a6)/(b5-b6), adding the IDM in CDIDN-v1 significantly enhances LLDs in DFs (around lesions and lung edges), while also maintaining smoothness in non-significant deformation areas of the lung data. This confirms that the IDM helps enhance LLDs in the DF, improving the model's DIC against LLDs in lungs. Meanwhile, enhancing differences between lesion locations and background plate promotes their separation.

The Jacobian determinant (J) intuitively reflects grid expansion and contraction, but it is limited by deformations in the DF, hindering identification of isolated small early lesions, which poses a risk. Intensity differences (IDCs) between fixed and warped images highlight underlying lesion disparities, especially for small early lesions, but minor IDCs of uniformly-intense lesions may obscure changes and mutilate lesions. In summary, we propose the $JAC\_Intensity$ metric as follows:
\begin{equation}
{JAC\_Intensity}=\frac{J-1}{ABS(J-1)} \times {(ABS(J-1)+N(C^{\Delta Its}))}^2,
\label{f9}
\end{equation}
\begin{figure}%[!htbp]
	\centering
	\includegraphics[width=1\textwidth]{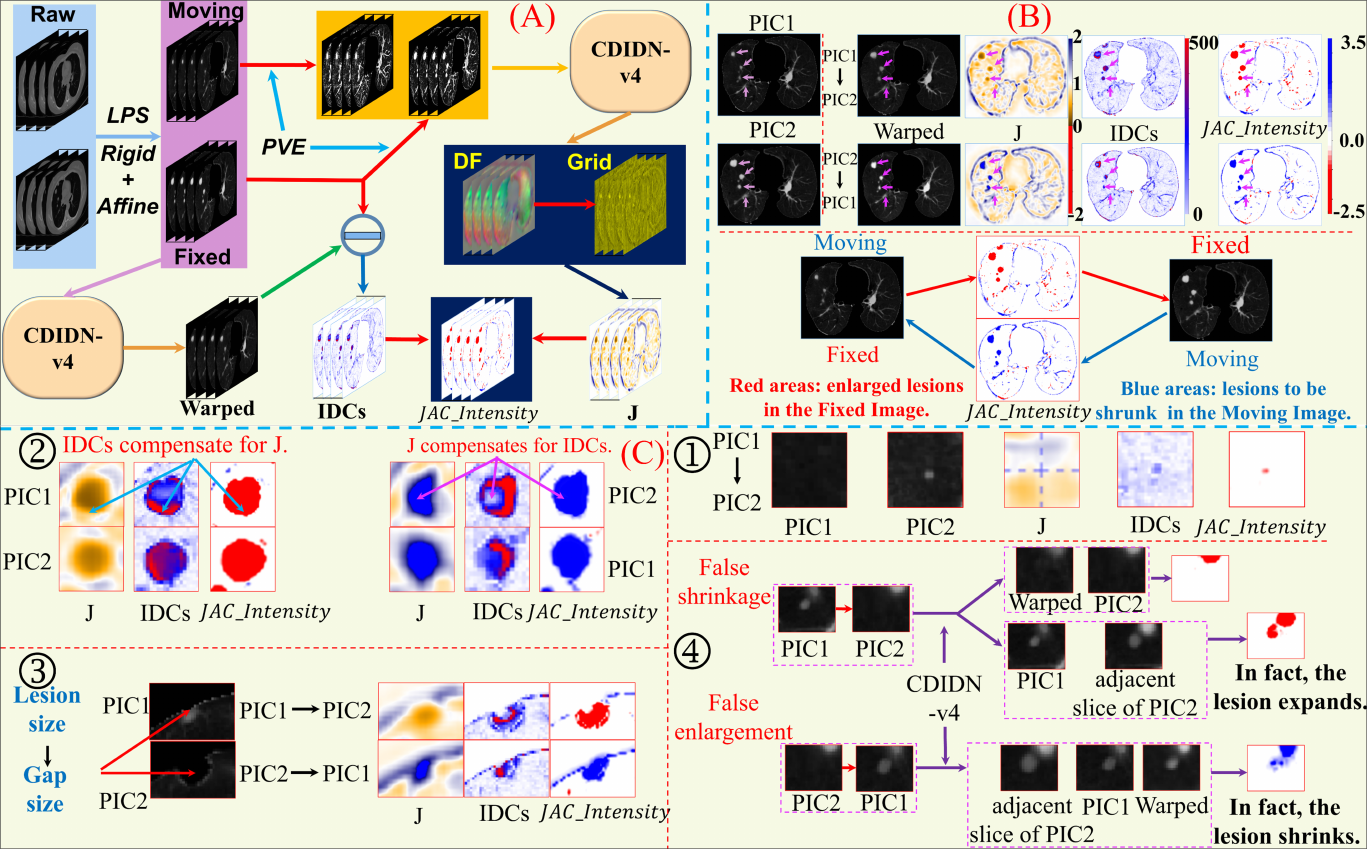} % 插入图片
	\caption{Long-term tracking of lesion dynamics. (A) Workflow. (B) Tracking results by $JAC\_Intensity$.  Arrows indicate key lesions. (C) Tracking effect in four conditions.}
	\label{p7}
\end{figure}
where $N(\cdot)$ denotes the normalization, $ABS(\cdot)$ denotes the absolute value operation. $C$ is a constant set empirically to 10, which exponentially accentuates the distinction between large IDCs and surrounding IDCs, thereby identifying the lesion locations with greater specificity. First, we truncate IDCs  as $\Delta Its$ using a window of [0, 120]. Then, we can calculate $JAC\_Intensity$ by equation \eqref{f9}. Figure~\ref{p7} (A) shows the workflow for long-term tracking of lesion dynamics using CDIDN-v4. Figure~\ref{p7} (B) shows tracking of lesion dynamics using $JAC\_Intensity$. Red regions mainly indicate increased or new lung lesions in later CT scans. Blue regions mainly indicate lesions in early scans that will shrink or disappear in later CT scans. Figure VII (C) shows tracking performance in four conditions. Condition 1 reflects monitoring of early isolated lesions  (1 mm or larger) using IDCs to compensate for J's deficiency. Condition 2 highlights contribution of J and IDCs to the integrity of lesions. The method uses J's high cohesiveness or expansiveness in lesion center to solve mutilation in the lesion center, and uses IDCs to compensate for mutilation of the lesion edges. Condition 3 reflects tracking dynamics of pulmonary marginal lesions while accounting for the problem of missing marginal lesions due to LPS. Tracking lesions is transformed into tracking gaps to reflect marginal lension dynamics. Condition 4 demonstrates our method's effectiveness at handling false shrinkage and enlargement of lung lesions. Lesions on a single slice can appear to change size visually, leading to incorrect interpretations of changes. By using our method to perform cross-slice alignment of LLD regions, we can avoid false judgments of shrinkage or enlargement post-alignment.

\section{Conclusions and Discussions}
In this paper, we propose CDIDN, which achieves excellent registration stability and accuracy for regions with LLD. The IDM in CDIDN significantly enhances model's DIC by improving its resilience to LLDs in CT images through enhancing LLDs in the DF. Leveraging this enhancement, we propose a method for tracking of lung lesion dynamics. To better reflect lesion changes, we generate the DF with enhanced LLDs by IDM on images after PVE, and generate IDCs on unenhanced images to avoid losing some lesion details due to the PVE. Limitations exist as it may show areas of high respiratory motion near lung edges despite the absence of lesions. We propose extending the application of IDM to other cascade- or feature-based methods to improve their DIC. Additionally, for both static and dynamic organs, we should focus more on the model's ability to register LLD regions, which are often the focus in medical practice. CDIDN has promising potential of addressing registration-based medical issues of other large-deformation organs or tissues, including the heart, stomach, bladder, and muscles, thus creating an exciting avenue for future work.

\section*{Acknowledgments and Disclosure of Funding}

Our work is supported by the National Key Research and Development Program of China (No.2021YFF1201200, No.20230201083GX), and the Science and Technology Development Program of Jilin Province, China (No.20210204100YY).

%\newpage
\bibliographystyle{plain}
\bibliography{neurips_2023.bib}

\end{document}